\RequirePackage{fix-cm}
\documentclass[twocolumn]{svjour3}          
\RequirePackage{rotating}
\usepackage{natbib}

\smartqed  
\usepackage{graphicx}
\usepackage[table,xcdraw]{xcolor}
\usepackage{lscape}
\usepackage{adjustbox}
\usepackage[hyphens]{url}
\usepackage{hyperref}
\usepackage{amsmath}
\journalname{}
\begin{document}

\title{Road obstacles positional and dynamic features extraction combining object detection, stereo disparity maps and optical flow data\thanks{This study was financed in part by the Coordenação de Aperfeiçoamento de Pessoal de Nível Superior - Brasil (CAPES) - Finance Code 001. CAPES (Brazilian Federal Agency for Support and Evaluation of Graduate Education). It was also supported by the Brazilian National Institute for Digital Convergence (INCoD), a research unit of the Brazilian National Institutes for Science and Technology Program (INCT) of the Brazilian National Council for Science and Technology (CNPq).}
}

\titlerunning{Road obstacles positional and dynamic features extraction}        

\author{Thiago Rateke\textsuperscript{1,2}         \and
        Aldo von Wangenheim\textsuperscript{1,2} 
}


\institute{Thiago Rateke \at
              \email{thiago@incod.ufsc.br}           
           \and
           Aldo von Wangenheim \at
              \email{aldo.vw@ufsc.br} 
           \and
          1 - Graduate Program in Computer Science, Federal University of Santa Catarina (PPGCC) - Department of Informatics and Statistics - Florianopolis, SC, Brazil \\
          2 - Image Processing and Computer Graphics Lab (LAPIX) at National Institute for Digital Convergence (INCoD)
}

\date{Received: date / Accepted: date}

\maketitle

\begin{abstract}
One of the most relevant tasks in an intelligent vehicle navigation system is the detection of obstacles. It is important that a visual perception system for navigation purposes identifies obstacles, and it is also important that this system can extract essential information that may influence the vehicle's behavior, whether it will be generating an alert for a human driver or guide an autonomous vehicle in order to be able to make its driving decisions. In this paper we present an approach for the identification of obstacles and extraction of class, position, depth and motion information from these objects that employs data gained exclusively from passive vision. We performed our experiments on two different data-sets and the results obtained shown a good efficacy from the use of depth and motion patterns to assess the obstacles' potential threat status.

\keywords{Features extraction \and Disparity Map \and Optical Flow}

\end{abstract}

\section{Introduction}
\label{sec:intro}
Obstacles detection in autonomous and drive-assisted vehicles concerns the detection of any other objects, static or in movement, on or near the road. In an intelligent autonomous vehicle navigation scenario it is, along with path detection, one of the most important tasks, because it involves not only the safety of the vehicle where the obstacles detection and recognition are performed, but also because it affects other participants in this scenario, such as: other vehicles, pedestrians, cyclists and animals. Based upon information continuously gathered by the obstacle detection, the behavior of an autonomous vehicle must adjust itself or, in the case of an Advanced Driver Assistance Systems (ADAS), it must generate alerts that allows drivers to adapt their driving to potential threats.

The state-of-the-art for obstacle detection is already quite robust, and with the recent advancements in convolutional neural network (CNN)-based deep learning approaches, has been obtaining excellent results. Prior to this work, we performed a Systematic Literature Review (\cite{rateke:2018} and \cite{rateke:2020}), based on the procedures described in \cite{kitchenham:2007}. And based on this literature review we were able to determine that the state-of-the-art in the road obstacle detection area with focus on vehicular navigation has many examples and different approaches.

The approaches we were able to identify vary widely. Some examples are: using only Stereo Vision (eg.: \cite{hane:2017}), only Optical Flow (eg.: \cite{bouchafa:2011}), Image Segmentation (eg.: \cite{poddar:2015}), and recently works using Convolutional Neural Networks (eg.: \cite{prabhakar:2017}). There are also several other approaches that use combinations between different methods, such as: \cite{gupta:2017} which uses methods based on Neural Networks, Stereo Vision and Image Segmentation, and \cite{giosan:2014} which uses methods based upon Stereo Vision, Optical Flow and Image Segmentation.

In moving obstacles detection, however, it is not only important to identify the kind of obstacle: other vehicle, a cyclist, a horse-rider, a pedestrian crossing the road or a stray animal. It is also important to be able to determine the potential path of these obstacles and be able to estimate if there exists the possibility of a collision, \emph{i.e.}, if each detected object has the potential to become a threat to the vehicle. For this purpose, it is necessary to be able to estimate the obstacles distance, velocity and direction of movement. In this context, we can understand Autonomous Vehicle Threat Assessment (AVTA) as the continuous active inspection of its sensorial data by a vehicle in order to identify road objects and traffic participants that could pose a threat to the vehicle's navigation.

In the  work we present here, we move on from obstacle detection to the next step, which is to identify, in the detected obstacles, features that are relevant to threat assessment in a vehicular navigation system context, such as: distance, velocity and also direction of movement from detected objects.

\subsection{Objectives}
The objective of this work is to investigate the feasibility of the development of a passive vision (PV)-based integrated moving obstacles detection and description approach that fulfills the following requirements:
\begin{itemize}
    \item detects and classifies obstacles pertaining to a set of predefined classes;
    \item provides depth information about each obstacle, relative to the vehicle;
    \item provides information about the trajectory and speed of each obstacle, relative to the vehicle;
    \item is capable of determining this information  employing only data gained from passive vision, without relying on additional data from LIDAR (Light Detection and Ranging) or other active sensors.
\end{itemize}

Furthermore, our work concentrated not in developing new image processing algorithms, but investigated if there exist already developed and mature technologies which could be combined in order to achieve the objective above. 

\subsection{Approach Outline}
In our approach, in the obstacle detection step, we employ stereo images and a state-of-the-art CNN structure, the Mask R-CNN \cite{he:2017}, which in addition to the detection and recognition of objects, also determines the position and shape of these objects, providing, as a second layer of results, a semantic segmentation (SS) of the recognized objects. From the original images, obtained from data-sets that provide stereo data with two-camera captures, we also generate the Disparity Maps (DM) of the scene (depth map) for each pair of stereo frames. This DM we apply to the objects recognized and segmented by the Mask R-CNN, extracting the average depth information for each segmented object, allowing spatial localization of these objects. In addition, we also apply the Optical Flow calculation on these objects, being able to filter the average movement flows (motion direction and intensity) separately for each detected object.

\subsection{Research Rationale}
Different sensors can be used for the obstacle detection task. Some vehicles employ an ensemble of diverse sensors, not only cameras \cite{urmson:2008}, and \cite{fernandes:2014}. One of these sensor, present in many autonomous vehicle navigation projects, are active sensors named as LIDARs, which are laser sources used for active sensing of reflected light, in order to measure distances between the sensor and the target object \cite{lidar:2015}. In vehicular projects, the LIDAR employed is normally a laser of \emph{Class1}, which is the category considered to present less danger. It employs light in the infra-red (IR) spectrum, in wavelengths in the order of 905nm. 

Based on the studies of \cite{slp:2015} and \cite{ans:2005}, a single \emph{Class1} laser source poses no danger to the retina when it does not remain in direct contact with the human eye for a longer time. A categorization of the lasers and the possible damages caused by excessive exposure in different levels is presented in \cite{slp:2015}. Lasers that emit in a wavelength between 780nm and 1400nm can cause cataracts and burn the retina. Considering a scenario where autonomous vehicles are used on a large scale, situations of dense traffic could be responsible for a many-LIDAR-originated ``lasersmog'' and become a risk to the nearest humans, which would simultaneously be targeted by the signals of many laser sources.

Even if there exists no conclusive study of the impact of many-car generated lasersmog on pedestrians yet, we understand that stereo camera-based PV may be a better alternative in a future autonomous vehicle scenario. For this purpose our work focuses on data achieved through passive stereo vision only, without any information supplementation through LIDAR data.

The remainder of this paper is organized as follows: In Section \ref{sec:relwor} we present the related works and their respective approaches. In Section \ref{sec:matmet} we present the data-sets used in our experiments and also present the methods we apply in our approach. Our approach is presented in Section \ref{sec:ourapp}. Followed by the results obtained in Section \ref{sec:results}. Finally, in Section \ref{sec:conclusions} we conclude this paper with a discussion about the results and the next steps in future work.

\section{Related work}
\label{sec:relwor}
Other authors have already tackled the PV-based extraction of relevant features from objects in the scene. \cite{mitzel:2011} performs a pedestrian detection with focus on multiple pedestrians tracking and uses Stereo Vision techniques for the detection step and the RANSAC framework for the pedestrian motion estimation and tracking.

In \cite{chanawangsa:2013} the authors present an approach to do the tracking of detected vehicles in the scene with focus on identifying overtaking situations. The markings on the road, the lanes, are also detected to know when a vehicle may be entering in front, allowing to generate an alert. For the vehicles detection step are used Histogram of Oriented Gradients (HOG) and Support Vector Machine classifier (SVM). The Kalman filter is used for vehicle tracking step.

An obstacle detection that also employs Stereo Vision techniques is presented in \cite{huang:2016}. Based on the image generated by the DM, the object contours are found. Based on these contours the authors use the objects' geometric information, such as area and height to classify objects (e.g.: people, vehicles and others). 

Other works also use geometric information from the detected obstacles in order to classify the obstacles by types. In \cite{li:2014}, the authors present an approach that besides the geometric information of the detected obstacles (height and width) uses fuzzy logic to classify these obstacles. In \cite{liu:2014} the authors applied a segmentation in the Disparity Map and also use width and height features from the detected obstacles to make the classification.

Stereo Disparity map is also used in \cite{chen:2012}) together with Histogram of Oriented Gradient (HOG) to extract the obstacles features. Finally, the classification of obstacles is made through a Support Vector Machine.

To predict future vehicle localization the authors from \cite{yao:2018} use a recurrent neural network (RNN) with a dense optical flow incorporation. In \cite{deo:2018} the authors also shown how to prevent other vehicles actions using hidden Markov model (HMM), interacting multiple model (IMM) and variational Gaussian mixture models (VGMM). Also to predict the trajectories from other vehicles the \cite{jawed:2018} presents an approach wich use a Convolutional neural network (CNN).

These works used different combinations of methods and techniques and are focused on classifying the types of obstacles or, at most, tracking some of the obstacles. Not focusing on extracting behavioral features from obstacles in relation to the moving vehicle.

\section{Material and methods}
\label{sec:matmet}
We employed two different data-sets in our experiments, both presenting stereo images from urban vehicle navigation scenarios, but in different contexts (Germany and Brazil). Both data-sets are presented in Section \ref{sec:datasets}. In Section \ref{sec:maskrcnn}, \ref{sec:dispmap} and \ref{sec:opticalflow} we provide a brief descriptions about each method we used in our model.

\subsection{Data-sets}
\label{sec:datasets}
The two data-sets used in our experiments were: KITTI data-set\footnote{http://www.cvlibs.net/datasets/kitti/raw\_data.php} \cite{geiger:2013} and CaRINA data-set\footnote{http://www.lrm.icmc.usp.br/dataset} \cite{shinzato:2016}. Both provide high-quality stereo images in vehicle navigation scenarios. KITTI uses a PointGray Flea2 cameras and CaRINA uses a Bumblebee XB3 camera.

Created by the Mobile Robot Laboratory group (ICMC / USP - Sao Carlos) filmed in Brazil, more specifically in the city of São Carlos in São Paulo state, the CaRINA data-set aimed to provide images for experiments in autonomous navigation visual perception in emerging countries scenarios, containing low quality roads. There are few pedestrian situations (almost none), but contains other vehicles in the scene (eg.: cars, motorbikes, trucks).

In contrast, the data-set provided by KITTI contains a considerable amount of pedestrian and cyclist situations in the scene, in addition to other vehicles. KITTI was created by the Karlsruhe Institute of Technology in Karlsruhe city, Germany. It is probably one of the most commonly used data-sets in visual perception works for vehicle navigation tasks, including for path detection and obstacle detection.

\subsection{Mask R-CNN}
\label{sec:maskrcnn}
\begin{figure}[!htb]
\centering
\includegraphics[width=\linewidth]{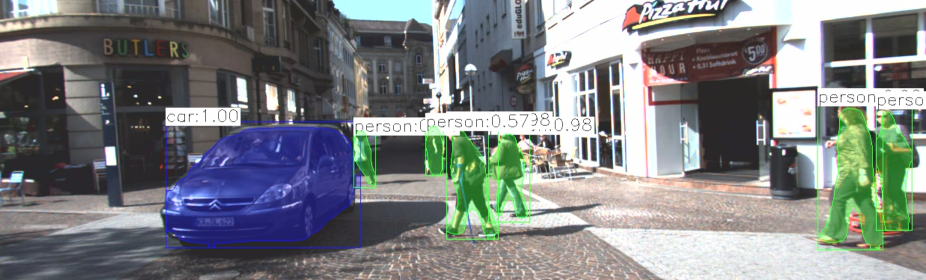}
\caption{Mask R-CNN example on KITTI data-set.}
\label{fig:maskrcnn}
\end{figure}

In \cite{he:2017} the authors present a framework for object instance segmentation. The Mask R-CNN, in addition to detecting and classifying objects in the scene, also applies a segmentation mask to each detected object (eg.: Figure \ref{fig:maskrcnn}). According to the authors, Mask R-CNN is an extension of Faster R-CNN \cite{ren:2015}.

We use this framework with pre-trained models in the Inception backbone architecture \cite{szegedy:2016}, which has good classification accuracy and is faster than many other architectures. Also, our experiments runs in a model trained with MSCOCO data-set \cite{yi:2014}, which is a data-set specific for object detection and segmentation.

\subsection{Disparity map}
\label{sec:dispmap}
The disparity is the difference that the same pixel has between two images, this difference takes into account the position of the same pixel in each images. It is common to use disparity as a synonym of depth \cite{bleyer:2013}. The ideal for Stereo Vision works is that the images are perfectly rectified on the \textit{y}-axis, allowing the scanning by checking the corresponding pixels and their respective differences to occur only on the \textit{x}-axis:

\begin{equation} 
\label{eq:disp}
D = x_{l} - x_{r}
\end{equation}

where $x_{l}$ is the specific pixel coordinate in left image, $x_{r}$ is the coordinate of the same specific pixel in the right image and \textit{D} is the disparity value between these points. Both data-sets used in our experiments have perfectly rectified images.

\begin{figure}[!htb]
\centering
\includegraphics[width=\linewidth]{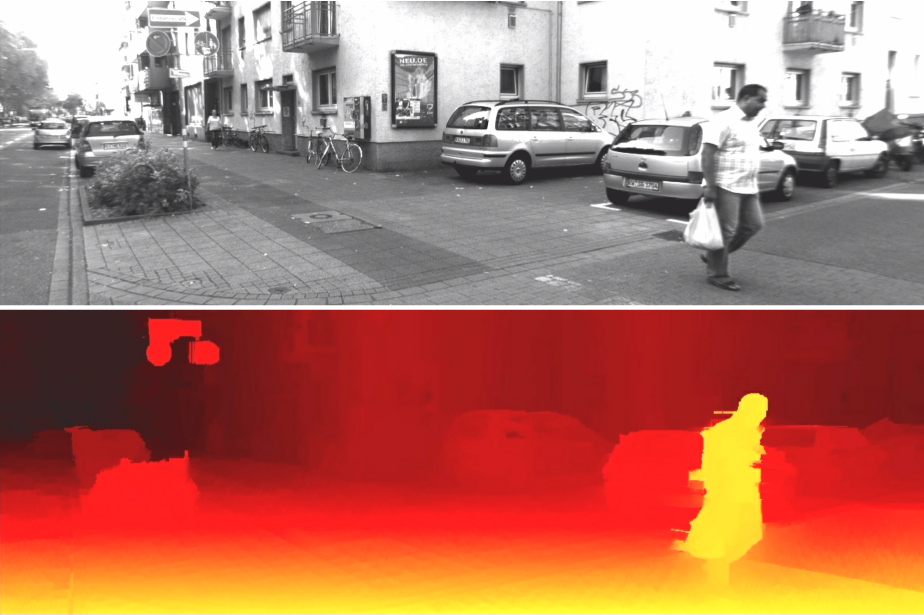}
\caption{Disparity Map example on KITTI data-set. Original left image on top, Disparity Map on bottom.}
\label{fig:disparitymap}
\end{figure}

The Disparity Map is the image that represents the pixel disparity values as an intensity image, where high intensity values represent high disparities and low intensity values represent lower disparities \cite{bleyer:2013}. Normally the Disparity Map is displayed as grayscale image, we applied a simple color conversion for a better visualization, but the intensity information is the same (eg.: Figure \ref{fig:disparitymap}).

\subsection{Optical flow}
\label{sec:opticalflow}
The goal of Optical Flow (OF) is to identify the displacement of intensity patterns in the image along sequential frames. This movement information can be very useful in computer vision because it also allows the identification of certain patterns in the scene \cite{fleet:2005}. 

In the literature there are examples of OF obtained through Neural Networks \cite{dosovitskiy:2015}\cite{ilg:2017} and also through traditional numeric methods \cite{farneback:2003}\cite{lucas:1981}. Neural OF methods may be a more recent tendency, but they also require more computational resources. In our work, we already performed the detection and segmentation of the obstacles with the use of a CNN and we only need to apply the OF calculations to the detected objects. For this reason we opted to perform a post-processing employing a traditional OF approach. In addition, this approach provides us with explicit vector data which could be later used by a vehicle for threat assessment, which is not possible with the present CNN-based OF approaches.

\begin{figure}[!htb]
\centering
\includegraphics[width=\linewidth]{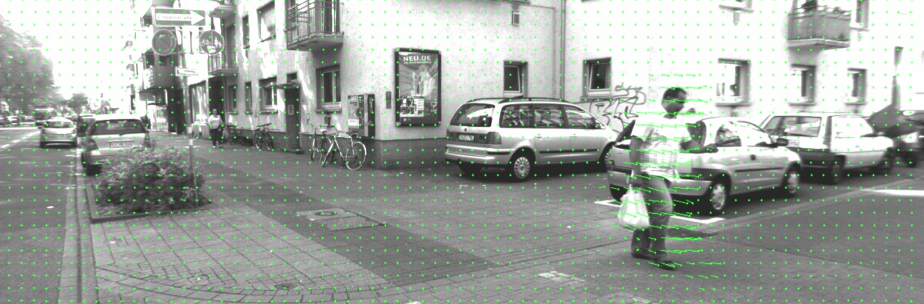}
\caption{Optical Flow example on KITTI data-set.}
\label{fig:opticalflow}
\end{figure}

In our approach we used the Gunnar-Farneback algorithm \cite{farneback:2003}, which produces a dense OF working on a grid of points. In this algorithm, the movement vector value is extracted through information obtained from two consecutive frames \cite{farneback:2003}. As this algorithm calculates the OF for each pixel in the image, it performs a good motion estimation of the regions encompassing the detected objects. An example with flow vectors is shown in Figure \ref{fig:opticalflow}, where a pedestrian is crossing the street in front of an awaiting vehicle.

\section{Our approach}
\label{sec:ourapp}
\begin{figure}[!htb]
\centering
\includegraphics[width=\linewidth]{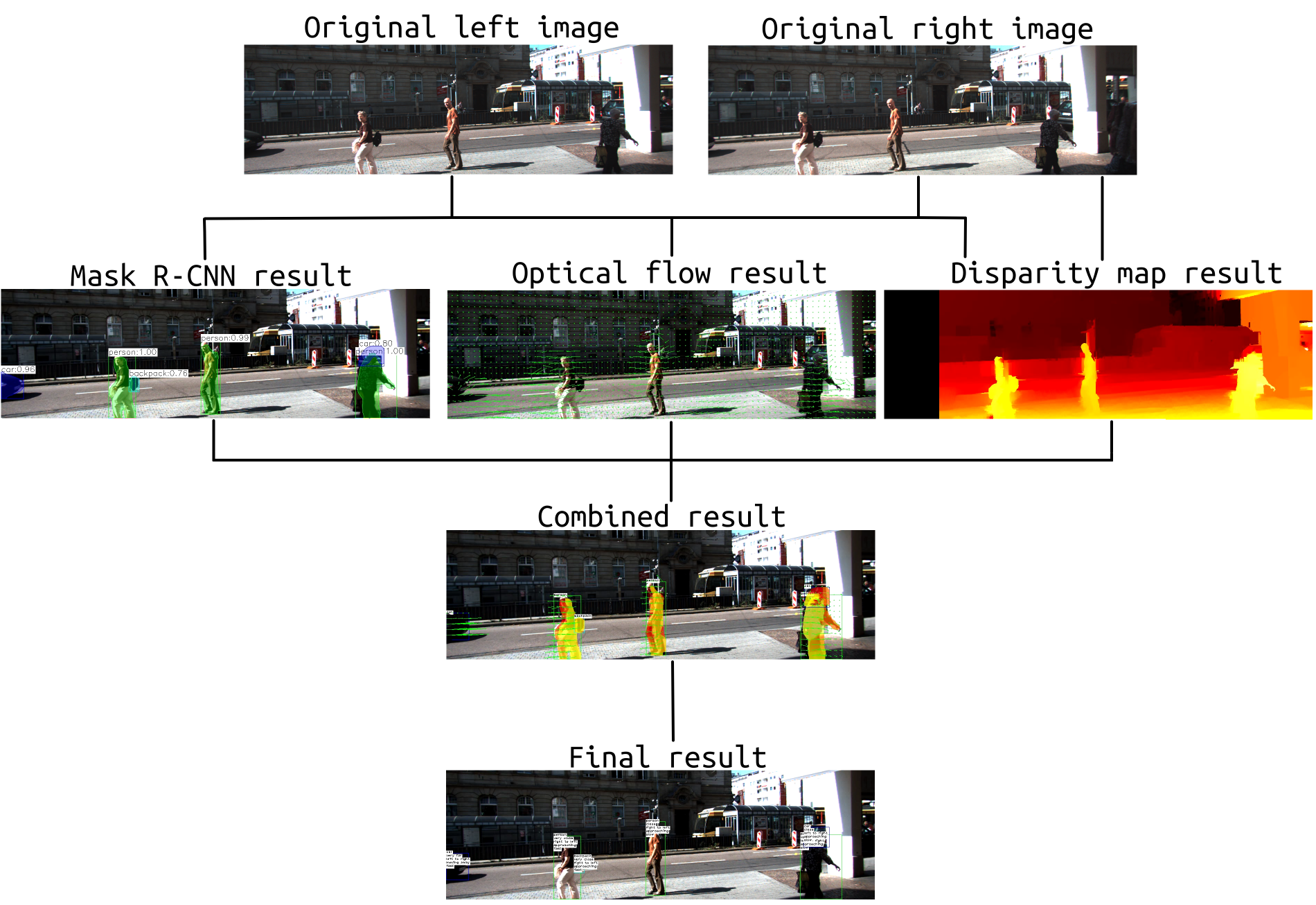}
\caption{The steps in our model.}
\label{fig:ourapproachmodel}
\end{figure}

In Figure \ref{fig:ourapproachmodel} we present a schematic overview of our approach, which consists in combining the techniques described in Section \ref{sec:matmet}. We integrate the obstacle detection and SS results obtained by Mask R-CNN with the results from OF and the Disparity Map. In this way, it is possible to extract the OF and disparity values from specific pixels in each object, which allows us to generate a detailed analysis of depth and movement for each object in the scene.

Both KITTI and CaRINA data-sets provide stereo images. For the OF calculation and the CNN object detection we employ only the left-captured (driver-side) images from the data-sets. For the calculation of the disparity maps we employ the whole stereo data.

With the disparity values obtained from each object segmentation, it is possible to generate an average disparity value for each segmented object. Thus, in the final analysis, we defined some depth labels with pre-set thresholds. We defined four depth labels: \textbf{very close}, \textbf{close}, \textbf{far} and \textbf{very far}.

In the same way, we used the Optical Flow values from each detected object to generate average movement values, computing their means as the resulting direction and intensity motion vector. We collected the direction values from the OF vector on the \textit{x}-axis to label whether the vehicle is stationary, or going from right to left or from left to right. We defined the direction labels on the \textit{x}-axis as: \textbf{left to right}, \textbf{right to left} and \textbf{stable direction}.

The direction values from the OF vector on the \textit{y}-axis indicate to us whether the vehicle is approaching, moving away or maintaining a stable distance. We defined three labels for the \textit{y}-axis being: \textbf{approaching}, \textbf{moving away} and \textbf{stable distance}.

The greater the displacement of a pixel between two frames, the greater will be the vector representing that displacement. This value allows us to have a sense whether the detected object is moving fast or slow. We obtained the displacement value from each object by multiplying the average values of the \textit{x}-axis and \textit{y}-axis from each object:

\begin{equation} 
\label{eq:flow01}
xM={\frac {1}{n}}\sum _{i=1}^{n}x_{i}
\end{equation}

\begin{equation} 
\label{eq:flow02}
yM={\frac {1}{n}}\sum _{i=1}^{n}y_{i}
\end{equation}

\begin{equation} 
\label{eq:flow03}
VL = xM * yM
\end{equation}

where \textit{xM} is the \textit{x}-axis average value in an object, \textit{yM} is the \textit{y}-axis average value in the same object, and \textit{VL} is the vector intensity value from that object. We defined five labels to represent the movement intensity: \textbf{stopped}, \textbf{slow}, \textbf{average speed}, \textbf{fast} and \textbf{very fast}.

\section{Results}
\label{sec:results}
We compared the results obtained with manual annotations made in a total of 415 obstacles over 100 frames, being part from the CaRINA dataset and part from the KITTI dataset. 20 sequences of frames were selected containing 5 frames each sequence. In Table \ref{tbl:accRes} it is presented the general accuracy for each task in the extraction and analysis of the obstacles positioning and movement. A more individual analysis is possible by the Tables \ref{tbl:confDepth}, \ref{tbl:confX}, \ref{tbl:confY} and \ref{tbl:confMov} where we presented, through the confusion matrices, the detailed results of each task and their labels.

\begin{table}[]
\centering
\caption{Accuracy results.}
\label{tbl:accRes}
\begin{tabular}{lll}
\hline\noalign{\smallskip}
task & accuracy \\
\noalign{\smallskip}\hline\noalign{\smallskip}
Depth & 81.75\% \\
x-axis Direction & 89.51\% \\
y-axis Direction & 83.57\% \\
Movement Intensity & 80.96\% \\
\noalign{\smallskip}\hline
\end{tabular}
\end{table}

\begin{table}[]
\centering
\caption{Depth Confusion Matrix}
\label{tbl:confDepth}
\begin{adjustbox}{width=\columnwidth}
\begin{tabular}{|
>{\columncolor[HTML]{C0C0C0}}c |c|c|c|c|}
\hline
 &
  \cellcolor[HTML]{C0C0C0}\begin{tabular}[c]{@{}c@{}}very close\end{tabular} &
  \cellcolor[HTML]{C0C0C0}close &
  \cellcolor[HTML]{C0C0C0}far &
  \cellcolor[HTML]{C0C0C0}\begin{tabular}[c]{@{}c@{}}very far\end{tabular} \\ \hline
\begin{tabular}[c]{@{}c@{}}very close\end{tabular} & \textbf{72.82\%} & 17.48\%          & 9.71\%           & 0.00\%            \\ \hline
close                                                 & 1.00\%           & \textbf{86.00\%} & 6.00\%           & 7.00\%            \\ \hline
far                                                   & 0.00\%           & 0.00\%           & \textbf{67.00\%} & 33.00\%           \\ \hline
\begin{tabular}[c]{@{}c@{}}very far\end{tabular}   & 0.00\%           & 0.00\%           & 0.00\%           & \textbf{100.00\%} \\ \hline
\end{tabular}
\end{adjustbox}
\end{table}

\begin{table}[]
\centering
\caption{x-axis Direction Confusion Matrix}
\label{tbl:confX}
\begin{adjustbox}{width=\columnwidth}
\begin{tabular}{|
>{\columncolor[HTML]{C0C0C0}}c |c|c|c|}
\hline
 &
  \cellcolor[HTML]{C0C0C0}\begin{tabular}[c]{@{}c@{}}left to right\end{tabular} &
  \cellcolor[HTML]{C0C0C0}\begin{tabular}[c]{@{}c@{}}right to left\end{tabular} &
  \cellcolor[HTML]{C0C0C0}\begin{tabular}[c]{@{}c@{}}stable direction\end{tabular} \\ \hline
\begin{tabular}[c]{@{}c@{}}left to right\end{tabular}    & \textbf{86.90\%} & 8.28\%           & 4.83\%           \\ \hline
\begin{tabular}[c]{@{}c@{}}right to left\end{tabular}    & 5.11\%           & \textbf{92.05\%} & 2.84\%           \\ \hline
\begin{tabular}[c]{@{}c@{}}stable direction\end{tabular} & 11.24\%          & 0.00\%           & \textbf{88.76\%} \\ \hline
\end{tabular}
\end{adjustbox}
\end{table}

\begin{table}[]
\centering
\caption{y-axis Direction Confusion Matrix}
\label{tbl:confY}
\begin{adjustbox}{width=\columnwidth}
\begin{tabular}{|
>{\columncolor[HTML]{C0C0C0}}c |c|c|c|}
\hline
 &
  \cellcolor[HTML]{C0C0C0}approaching &
  \cellcolor[HTML]{C0C0C0}\begin{tabular}[c]{@{}c@{}}moving away\end{tabular} &
  \cellcolor[HTML]{C0C0C0}\begin{tabular}[c]{@{}c@{}}stable distance\end{tabular} \\ \hline
approaching                                                & \textbf{78.40\%} & 8.45\%           & 13.15\%          \\ \hline
\begin{tabular}[c]{@{}c@{}}moving away\end{tabular}     & 0.00\%           & \textbf{89.13\%} & 10.87\%          \\ \hline
\begin{tabular}[c]{@{}c@{}}stable distance\end{tabular} & 5.16\%           & 5.81\%           & \textbf{89.03\%} \\ \hline
\end{tabular}
\end{adjustbox}
\end{table}

\begin{table}[]
\centering
\caption{Movement Intensity Confusion Matrix}
\label{tbl:confMov}
\begin{adjustbox}{width=\columnwidth}
\begin{tabular}{|
>{\columncolor[HTML]{C0C0C0}}c |c|c|c|c|c|}
\hline
 &
  \cellcolor[HTML]{C0C0C0}stopped &
  \cellcolor[HTML]{C0C0C0}slow &
  \cellcolor[HTML]{C0C0C0}\begin{tabular}[c]{@{}c@{}}average \\ speed\end{tabular} &
  \cellcolor[HTML]{C0C0C0}fast &
  \cellcolor[HTML]{C0C0C0}\begin{tabular}[c]{@{}c@{}}very \\ fast\end{tabular} \\ \hline
stopped                                                  & \textbf{90.41\%} & 5.48\%           & 4.11\%           & 0.00\%           & 0.00\%           \\ \hline
slow                                                     & 13.79\%          & \textbf{58.62\%} & 24.14\%          & 0.00\%           & 0.00\%           \\ \hline
\begin{tabular}[c]{@{}c@{}}average \\ speed\end{tabular} & 0.00\%           & 1.49\%           & \textbf{77.61\%} & 20.90\%          & 0.00\%           \\ \hline
fast                                                     & 0.51\%           & 1.53\%           & 10.71\%          & \textbf{84.69\%} & 2.55\%           \\ \hline
\begin{tabular}[c]{@{}c@{}}very \\ fast\end{tabular}     & 0.00\%           & 0.00\%           & 4.00\%           & 26.00\%          & \textbf{70.00\%} \\ \hline
\end{tabular}
\end{adjustbox}
\end{table}

In Table \ref{tbl:confDepth}, the confusion matrix of the distance (depth) analysis of the obstacles in the scene, it is noticed that the worst results were with the labels ``very-close'' and ``far''. However, it is also possible to verify that the biggest errors in both classes were in neighboring labels. Still, the ``very-close'' label featured a considerable amount of errors as ``far''. This occurs in situations when there are objects close between each other in the scene making it difficult to analyze as individual objects. The same occurs with the ``approaching'' label in Table \ref{tbl:confY}, which has 8,45\% being as ``moving-away''. In Table \ref{tbl:confMov} the worst result was in label ``slow'', yet in all classes have errors occurring as being from neighboring labels.

\begin{figure*}[!htb]
    \centering 
	\includegraphics[width=1\linewidth]{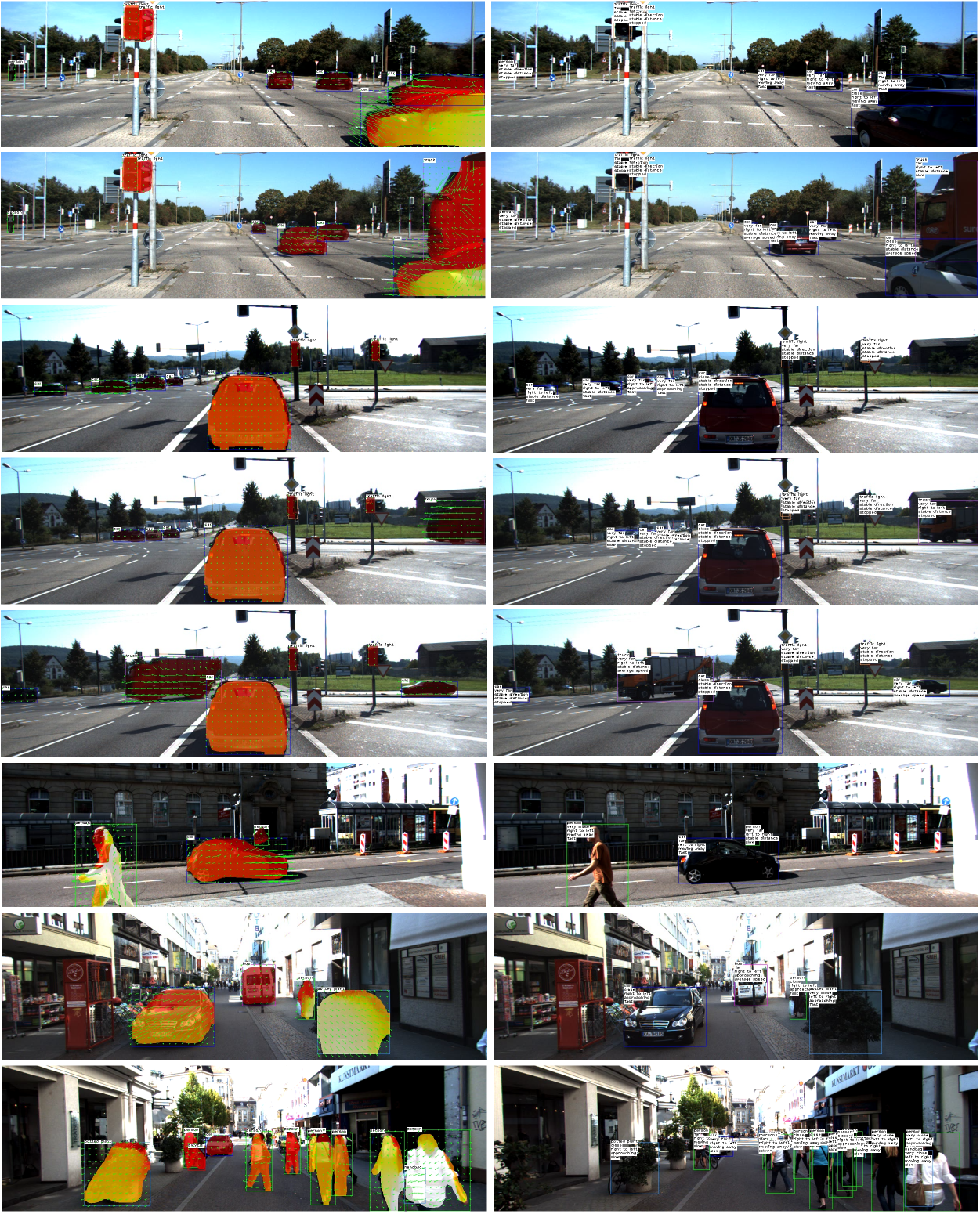}
	\caption{Examples of results in the KITTI data-set.}
	\label{fig:kittiresults}
\end{figure*}

\begin{figure*}[!htb]
    \centering 
	\includegraphics[width=1\linewidth]{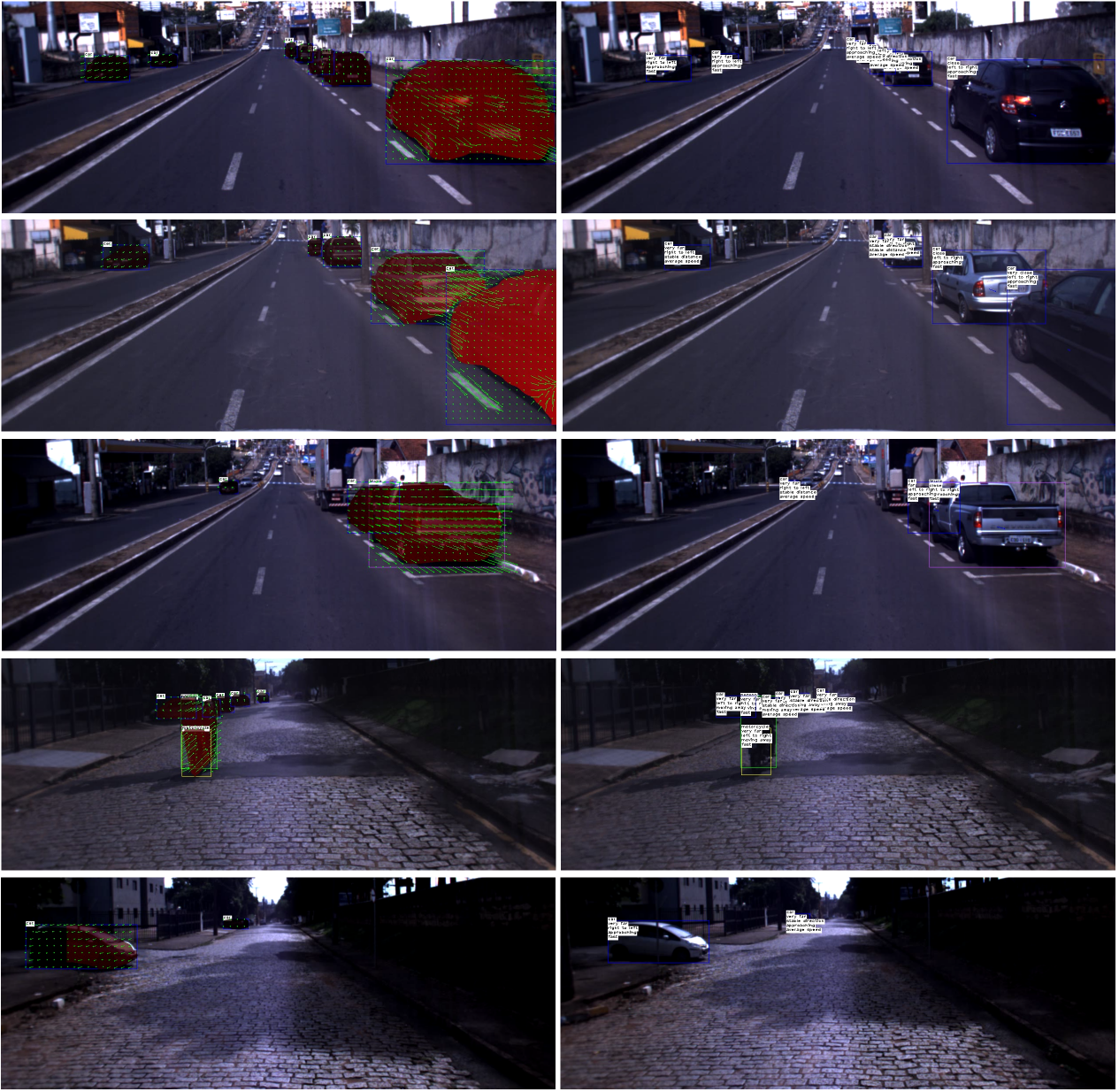}
	\caption{Examples of results in the CaRINA data-set.}
	\label{fig:carinaresults}
\end{figure*}

In Figure \ref{fig:kittiresults} and in Figure \ref{fig:carinaresults} we present some of the results obtained by our approach. In the left column we show the combined results obtained with the CNN, together with the Disparity Map and OF patterns. In the right column we present the labels on the objects based on the analysis of the depth and movement patterns of each object.

\begin{figure}[!htb]
\centering
\includegraphics[width=\linewidth]{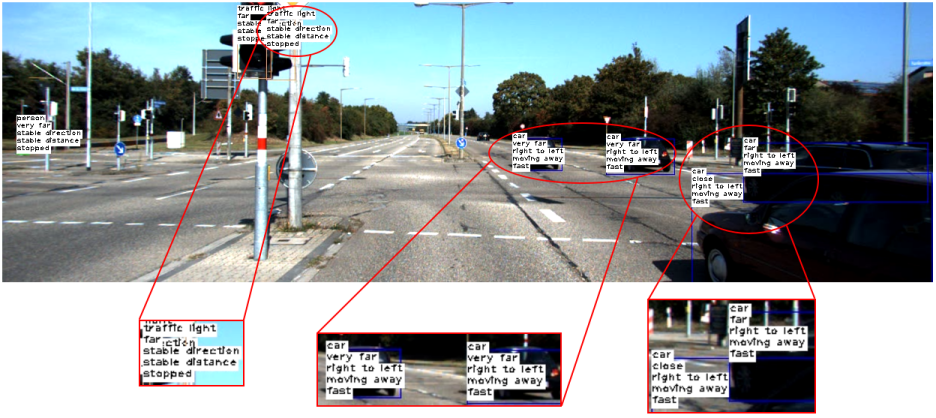}
\caption{Highlighted pattern analysis results.}
\label{fig:kittiresults_highlighted_01}
\end{figure}

In the first row of Figure \ref{fig:kittiresults} the vehicle responsible for the capture of the images (capture source, CS) is stationary, and four cars are passing through the right lane. Here, using the patterns obtained by Disparity Map and OF it is possible to verify their behavior and how distant these four vehicles are, as highlighted in Figure \ref{fig:kittiresults_highlighted_01}. In comparison, traffic lights are identified as static. The second row of Figure \ref{fig:kittiresults} presents the continuity of the first row, with the CS still stationary and vehicles passing through the right lane.

The vehicles passing through the right lane, present as a result of direction the label \textbf{right to left} because even though it is not crossing abruptly in front of the CS, it is not an exactly parallel movement because the image perspective, by surpassing the CS it is like moving in the \textit{x}-axis, from \textbf{right to left}. Considering this perspective, as lines going to meet at the vanishing point.

\begin{figure}[!htb]
\centering
\includegraphics[width=\linewidth]{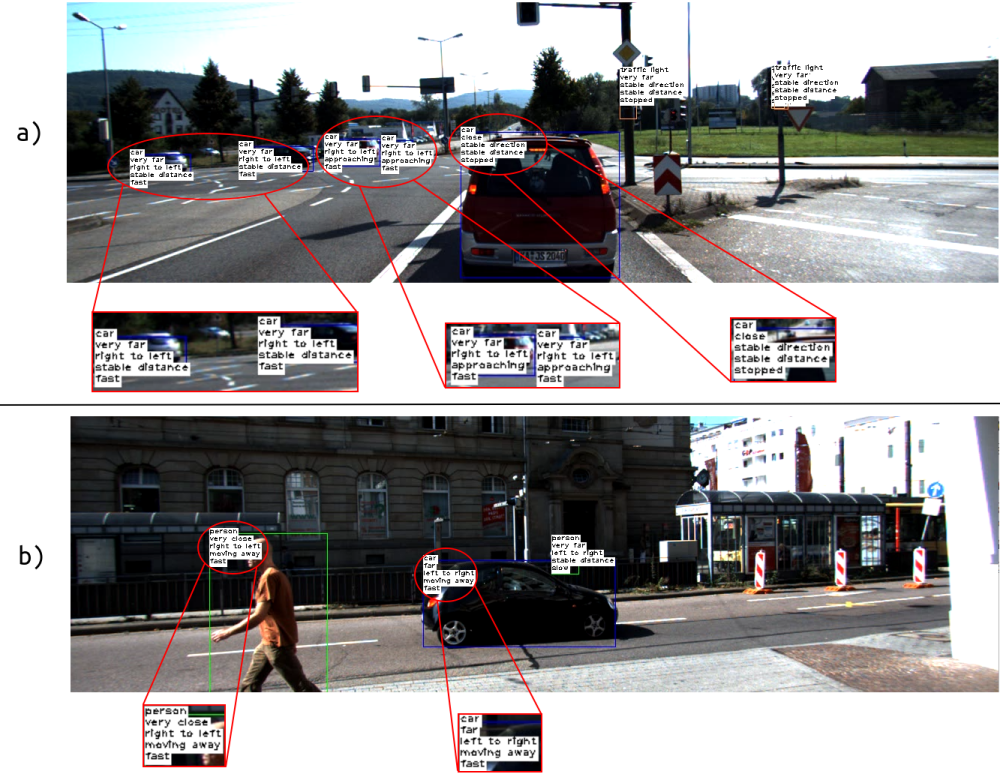}
\caption{Highlighted pattern analysis results.}
\label{fig:kittiresults_highlighted_02}
\end{figure}

The third, fourth and fifth rows of Figure \ref{fig:kittiresults} are a sequence and presents vehicles with a trajectory that will generate an actual direct crossing. These vehicles are further away while the CS is standing behind another, nearby vehicle, which is also stationary (Figure \ref{fig:kittiresults_highlighted_02} a).

The figures in the fourth and fifth rows of Figure \ref{fig:kittiresults} show a truck crossing the front of the vehicle from right to left. In the sixth row, we show the extraction and analysis of the patterns on a pedestrian very close to the CS and another vehicle more distant, both crossing the front of the CS in opposite directions (Figure \ref{fig:kittiresults_highlighted_02} b). Different objects with different distances are shown on the seventh row.

In the last row of Figure \ref{fig:kittiresults} we present a sequence where the CS is slowing down, almost stopping, while several pedestrians begin to cross with similar but not synchronized behavior, resulting in some data variation.

In the five rows of Figure \ref{fig:carinaresults}, showing results from CaRINA data-set, the CS is in motion and, although it presents images with less movement than in the KITTI data-set, it is still possible to observe the patterns of movement and distance from the detected objects. Mainly from first to third row, which correspond to a sequence.

\section{Conclusion and discussion}
\label{sec:conclusions}
Obstacle detection and recognition focused on ADAS and/or on Autonomous Vehicles navigation has made a major breakthrough in the state of the art in recent years, especially considering the advances in CNNs. The approach we present in this paper focuses on the next step after the detection and recognition of obstacles: the extraction of the depth and movement patterns of the detected objects.

We understand that identifying these patterns will allow a smarter and safer decision making in an ADAS or in an Autonomous Vehicle, helping to identify potential threats. Both providing for a more precise alert for a human driver, as well as passing more data to an intelligent agent module responsible for making decisions in an Autonomous Vehicle.

In our approach, we combine CNN-based detection and object recognition results with the depth patterns by a Disparity Map and movement patterns (direction and velocity) by an Optical Stream. The results obtained are promising and motivate the continuity of this research.

\subsection{Future work}
\label{sec:futureworks}
One of our next steps with this approach consists of applying that same model, with the same combined methods in a NVIDIA Jetson card, provided by NVIDIA for our project through the NVIDIA GPU Grant Program, in a vehicle with real time image capture and a specially developed stereo rig. Thus, improving the performance of the current proposed flow.

In addition, one of the possibilities opened by the extraction of the distance, trajectory and movement patterns we are performing, is to try to predict the actions of the participants in the scene, such as other vehicles, cyclists, pedestrians and animals, performing \emph{threat assessment}, which is a project that is already underway in our group.

In the context of these possible next steps we are also investigating the possibilities associated to the analysis of the obtained patterns, studying the creating of potential new behavior labels. During this next phases we plan to perform more experiments related to the obtained patterns analysis, differentiating for example the direction label in order to differentiate situations where in fact there will occur some crossing in front of the capture source from when it is a possible lateral overtaking.

%
\section*{Conflicts of interest}
The authors declare that there are no conflicts of interest.

\bibliographystyle{spbasic}      
\bibliography{refs}   

\end{document}